\documentclass[preprint,12pt]{elsarticle}

\usepackage{amssymb}
\usepackage{amsmath}

\usepackage{amsfonts}
\usepackage{algorithmic}
\usepackage{algorithm}
\usepackage{array}
\usepackage[caption=false,font=normalsize,labelfont=sf,textfont=sf]{subfig}
\usepackage{textcomp}
\usepackage{stfloats}
\usepackage{url}
\usepackage{verbatim}
\usepackage{graphicx}

\usepackage{graphicx}
\usepackage{bbm}
\usepackage{multirow}
\usepackage{xcolor}
\usepackage{wrapfig}
\usepackage{comment}
\usepackage{booktabs}
\usepackage{pifont}
\usepackage{makecell}
\usepackage{caption}

\journal{Pattern Recognition}

\begin{document}

\begin{frontmatter}



\title{Brain Foundation Models with Hypergraph Dynamic Adapter for Brain Disease Analysis}


\author[label1]{Zhongying Deng\corref{Corresponding}\fnref{fn1}}
\cortext[Corresponding]{Corresponding author. Email: zd294@cam.ac.uk}
\fntext[fn1]{Contributed equally.}
\affiliation[label1]{organization={Department of Applied Mathematics and Theoretical Physics, University of Cambridge},
            city={Cambridge},
            postcode={CB3 0WA}, 
            country={UK}}
            
\author[label2]{Haoyu Wang\fnref{fn1}}
\affiliation[label2]{organization={Shanghai Artificial Intelligence Laboratory},
            city={Shanghai},
            postcode={200232}, 
            country={China}}

\author[label2]{Ziyan Huang\fnref{fn1}}

\author[label1]{Lipei Zhang\fnref{fn1}}

\author[label3]{Angelica I.  Aviles-Rivero}
\affiliation[label3]{organization={ Yau Mathematical Sciences Center, Tsinghua University},
            city={Beijing},
            postcode={100084}, 
            country={China}}

\author[label1]{Chaoyu Liu}

\author[label2]{Junjun He}

\author[label4]{Zoe Kourtzi}
\affiliation[label4]{organization={Department of Psychology, University of Cambridge},
            city={Cambridge},
            postcode={CB2 3EB}, 
            country={UK}}
            
\author[label1]{Carola-Bibiane Schönlieb}

\begin{abstract}
Brain diseases, such as Alzheimer's disease and brain tumors, present profound challenges due to their complexity and societal impact. Recent advancements in brain foundation models have shown significant promise in addressing a range of brain-related tasks. However, current brain foundation models are limited by task and data homogeneity, restricted generalization beyond segmentation or classification, and inefficient adaptation to diverse clinical tasks. In this work, we propose SAM-Brain3D, a brain-specific foundation model trained on over 66,000 brain image-label pairs across 14 MRI sub-modalities, and Hypergraph Dynamic Adapter (HyDA), a lightweight adapter for efficient and effective downstream adaptation. SAM-Brain3D captures detailed brain-specific anatomical and modality priors for segmenting diverse brain targets and broader downstream tasks. HyDA leverages hypergraphs to fuse complementary multi-modal data and dynamically generate patient-specific convolutional kernels for multi-scale feature fusion and personalized patient-wise adaptation. Together, our framework excels across a broad spectrum of brain disease segmentation and classification tasks. Extensive experiments demonstrate that our method consistently outperforms existing state-of-the-art approaches, offering a new paradigm for brain disease analysis through multi-modal, multi-scale, and dynamic foundation modeling.

\end{abstract}

\begin{graphicalabstract}
\includegraphics[trim=40 10 210 10,clip,width=\linewidth]{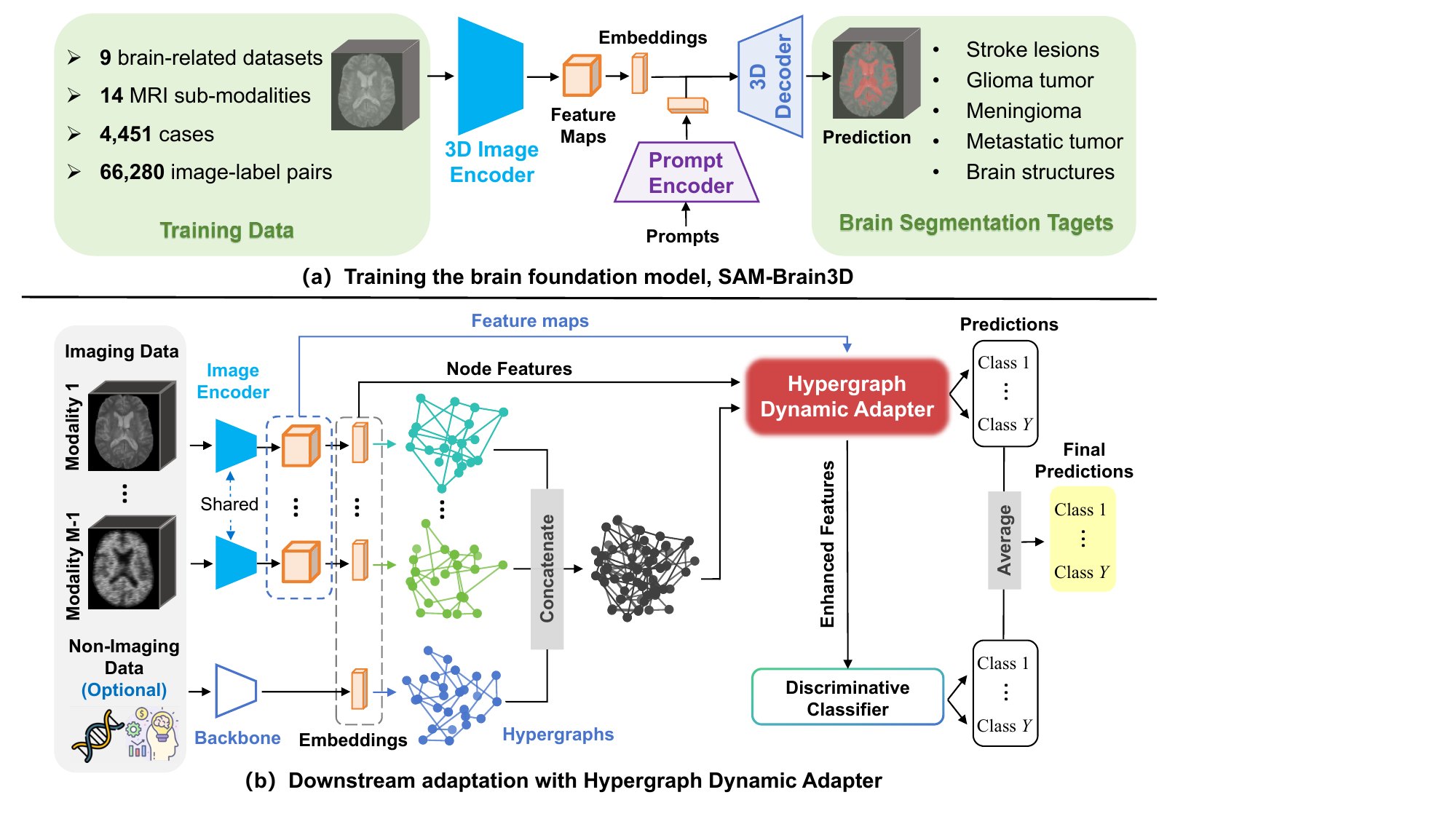}
\end{graphicalabstract}

\begin{highlights}
\item We introduce a brain foundation model termed SAM-Brain3D, which not only segments diverse seen and unseen brain targets, but also adapts to diverse downstream brain disease classification tasks, promoting strong generalization and transferability.

\item We propose a novel Hypergraph Dynamic Adapter (HyDA), enabling the encoder of SAM-Brain3D to adapt efficiently to heterogeneous input modalities via multi-scale and patient-specific dynamic adaptation across different tasks.

\item Our model consistently surpasses current state-of-the-art methods across a variety of brain disease analysis tasks.
\end{highlights}


\end{frontmatter}



\section{Introduction}
Brain-related diseases—such as Alzheimer’s disease and brain tumors—constitute some of the most significant challenges in contemporary medicine due to their profound individual and societal impacts \cite{jung2023conditional,cox2024brainsegfounder}. Nevertheless, the process of understanding and diagnosing these brain-related diseases is intrinsically complex, primarily due to the need to integrate highly heterogeneous data. Such data span both structural and functional neuroimaging modalities (e.g., MRI and PET), in addition to non-imaging clinical information such as genetic profiles, demographic data, and cognitive scores.

In recent years, artificial intelligence (AI) has become increasingly instrumental in facilitating disease diagnosis and prognosis, largely through its ability to process and analyze large-scale multi-modal medical data \cite{ma2024segment,liu2025vision}. 
To leverage these large-scale data, the medical AI community has begun to develop multi-modal foundation models tailored specifically to clinical domains \cite{ma2024segment,jiang2024neurolm,wang2024eegpt}. These foundation models has transformed this landscape by delivering impressive performance and generalizability across diverse medical tasks. 
Among them, models based on the Segment Anything Model (SAM)—such as MedSAM \cite{ma2024segment}, SAM-Med2D \cite{cheng2023sam}, and SAM-Med3D \cite{wang2023sam}—have achieved remarkable success in medical segmentation tasks. 

Despite these advances, existing SAM-based models are largely confined to segmentation and do not generalize effectively to other types of tasks. Moreover, general-purpose medical foundation models like SAM are typically trained on datasets involving non-neurological organs, which limits their ability to address brain-specific challenges. To overcome these limitations, several brain-focused foundation models have recently been introduced \cite{caro2024brainlm,tak2024foundation}. These models are specialized for brain-related tasks. However, they still suffer from task and data homogeneity: most are designed for a single type of tasks, e.g., either segmentation or classification \cite{wang2024eegpt}, and for handling a specific data modality, e.g., either EEG \cite{jiang2024large} or MRI \cite{cox2024brainsegfounder,wang2024eegpt}. This homogeneity restricts their versatility and broader applicability.

Another major limitation of current brain foundation models is that they lack a powerful mechanism for efficient and effective adaptation to diverse downstream tasks. Some models adopt a cumbersome dual-stage pre-training followed by a third stage fine-tuning \cite{cox2024brainsegfounder}, thus less efficient. Others either fail to handle both imaging and non-imaging modalities \cite{barbano2024anatomical}, or fall short in simultaneously achieving multi-scale fusion (i.e., fusing low-level morphological and high-level semantic features) and personalized patient-specific adaptation \cite{castellano2024automated}. These shortcomings significantly limit their performance, particularly given the need for complementary multi-modal information, multi-scale morphological and semantic features, and individualized patient analysis in brain disease diagnosis.

In this work, we present a new brain foundation model, SAM-Brain3D, and a Hypergraph Dynamic Adapter (HyDA) to address the above challenges. SAM-Brain3D is trained on 9 brain segmentation datasets, encompassing 14 MRI sub-modalities and 66,280 image-label pairs from 4,451 cases. This enables the model to capture detailed anatomical and modality-specific characteristics of the brain for segmenting diverse brain targets, further providing robust prior knowledge for diverse downstream tasks. HyDA is a lightweight adapter designed to efficiently and effectively adapt SAM-Brain3D to downstream brain disease analysis. HyDA leverages hypergraphs to extract complementary information from multi-modal data. It further exploits the high-level semantic features from the hypergraph to dynamically generate convolutional kernels for each patient, with the semantic kernels used to convolve with low-level features for multi-scale feature fusion. The dynamic patient-specific kernels further facilitate personalized adaptation for each individual patient. With SAM-Brain3D and HyDA, our method can excel in diverse clinical tasks, including both segmentation and classification tasks. 
Our main contributions are summarized as follows:

\begin{itemize}

\item \textbf{Brain Foundation Model for Diverse Tasks:} Unlike prior models that predominantly focus on a single task type, our SAM-Brain3D  not only segments brain tumor, meningioma, up to 35 brain structures, and broader unseen targets, but also adapts to diverse downstream brain disease classification tasks, promoting strong generalization and transferability.

\item \textbf{Multi-modal, Multi-scale, and Dynamic Adaptation:} We propose a novel Hypergraph Dynamic Adapter (HyDA), enabling the encoder of SAM-Brain3D to adapt efficiently to heterogeneous input modalities via multi-scale and patient-specific dynamic adaptation across different tasks.

\item \textbf{Superior Downstream Performance:} Our model consistently surpasses current state-of-the-art methods across a variety of brain-related downstream tasks.
\end{itemize}

\section{Related Work}
\paragraph{Brain Foundation Models}

Brain foundation models have emerged as a promising direction in AI for brain disease analysis. These models aim to learn comprehensive and transferable representations of brain anatomy and function from neuroimaging data. 
Some works have focused exclusively on brain MRI to develop specialized foundation models \cite{barbano2024anatomical,cox2024brainsegfounder,tak2024foundation} without adapting them to other modalities like PET and non-images.  In parallel, several studies focus solely on functional modalities such as fMRI and EEG to develop brain foundation models for capturing dynamic neural activity patterns \cite{jiang2024large,wang2024eegpt}.
In this work, we introduce a brain foundation model with a novel Hypergraph Dynamic Adapter (HyDA) to support a broader range of clinical and predictive tasks, e.g., brain segmentation and disease diagnosis. Our method can tackle both imaging and non-imaging modalities and facilitate multi-scale feature fusion using a dynamic convolution for better personalized patient-specific disease analysis. 

\paragraph{SAM for Medical Tasks}
Vision foundation models have achieved remarkable success in general-purpose visual understanding. Among them, the Segment Anything Model (SAM) \cite{kirillov2023segment} stands out for its zero-shot and promptable segmentation capabilities. Following its success on natural images, researchers have begun exploring SAM’s applicability to medical imaging.  Several studies have demonstrated its effectiveness on 2D medical image segmentation across various modalities \cite{cheng2023sam,ma2024segment},  such as MRI and CT scans. 
Recently, Wang et al. show that training a fully 3D SAM architecture from scratch on a large-scale medical dataset leads to superior segmentation performance and better generalization \cite{wang2023sam}. However, existing SAM-based medical foundation models remain predominantly focused on segmentation tasks, with limited extension to other predictive tasks such as classification. This narrow focus restricts their versatility and applicability. Moreover, SAM-based models are typically designed for general-purpose medical imaging tasks and lack specialization for brain-specific applications.
In this work, we propose a brain foundation model built upon a 3D SAM framework and trained on a large-scale multi-modal dataset encompassing both structural and functional imaging. Equipped with a Hypergraph Dynamic Adapter, our method can flexibly support various downstream tasks, including both segmentation and classification. Furthermore, it can accommodate tasks involving non-imaging clinical data including genetic profiles, demographic factors, and cognitive assessments.

\paragraph{Hypergraph Neural Networks}
Hypergraph Neural Networks (HNNs) \cite{feng2019hypergraph} have emerged as powerful tools for modeling complex, high-order relations in data. Unlike Graph Neural Networks (GNNs), which model pairwise relations, HNNs represent interactions among multiple nodes through hyperedges. This structure enables HNNs to capture more expressive and structurally rich dependencies \cite{gao2022hgnn}, making HNNs particularly well-suited for learning from multi-modal datasets where entities are often interrelated in intricate ways \cite{jing2025multi,yang2024hypercomplex}. 
In multi-modal learning, HNNs have demonstrated strong capabilities in integrating diverse data types. For example, Yang et al. \cite{yang2024hypercomplex} introduce a hyper-complex graph neural network that enables deep coupling across modalities through cross-embedding and aggregation. 
Inspired by these advances, we incorporate an HNN-based adapter, HyDA, into our brain foundation model to enhance its adaptability across tasks. HyDA allows the encoder to dynamically adjust to different data distributions including MRI, PET, and non-imaging clinical data, thereby supporting a broad range of downstream neuroimaging and clinical prediction tasks. By leveraging the expressive capacity of HNNs, our model enhances generalization and achieves superior task-specific performance across diverse clinical scenarios.

\paragraph{Alzheimer's Progression} 
Alzheimer's disease (AD) can lead to a decline in brain functions, such as memory loss and cognitive impairment. Alzheimer's progression aims to detect AD at the mild symptomatic or asymptomatic stage so that the early treatment can be enforced to slow down the progression of AD~\cite{lei2023multi}. Recent studies~\cite{huang2019diagnosis,shen2021heterogeneous} have focused on leveraging multi-modal data to extract complementary information for better Alzheimer's progression. 
Multi-modal CNN~\cite{huang2019diagnosis} simply concatenates features of MRI and PET modalities for multi-modal fusion. ProAuto-noAge~\cite{shen2021heterogeneous} resorts to auxiliary data of the structural MRI information of AD and Normal Control in addition to PET scans.  Multi-scale graph-based grading~\cite{hett2021multi} uses different graphs to process different anatomical scales (e.g., brain structures and hippocampal subfields), with multi-scale features fused for final prediction. It also enforces a consistent predictions from MRI and PET modalities to exploit multi-modal information. HAN~\cite{li2023early} employs hypergraph followed with attention to fuse MRI and Morphology modalities. VAP-Former~\cite{kang2023visual} leverages prompt fine-tuning to adapt a pre-trained transformer to MCI progression prediction task and utilizes a transformer block to integrate imaging and non-imaging modalities. The latest MMSDL~\cite{abdelaziz2025multi} from 2025 adopts different backbones to process MRI and PET images of different resolutions/scales which are fused via cross-attention. Our method differs from them in two folds. 1) We introduce a brain foundation model for segmenting diverse targets in brain images; 2) We propose a Hypergraph Dynamic Adapter using features from hypergraphs to dynamically generate convolutional kernels for convolution-based multi-scale feature fusion.

\paragraph{MGMT Classification}
Recent studies have explored the prediction of 6-methylguanine-DNA methyltransferase (MGMT) promoter methylation status in glioblastoma patients using deep learning applied to MRI scans. Early CNN-based models \cite{chang2018deep, korfiatis2017residual} report promising accuracies above $80\%$ on datasets from TCGA and institutional cohorts. Yogananda et al.\cite{yogananda2021mri} further achieve over $90\%$ accuracy using a T2WI-only 3D-DenseUNet. However, large-scale evaluations, such as the Brain Tumor Radiogenomic Classification Challenge \cite{baid2021rsna}, reveal the intrinsic difficulty of the task, where top solutions cannot exceed an AUC of 62\%. Later efforts~\cite{emchinov2021deep} confirm that directly training deep models on MRI data faces challenges in capturing complex imaging biomarkers. Traditional CNNs, while effective locally, are limited by their reliance on fixed receptive fields, restricting their ability to model global contextual information. To address this, graph-based approaches have been proposed. For instance, Hu et al.~\cite{hu2024mgmt} introduce a vision graph neural network (ViG) that represents MRI data as graphs, enabling the capture of global and irregular structures. This design alleviates the locality limitations of CNNs and improves performance in MGMT methylation prediction. Despite these advances, existing models still heavily rely on task-specific training with limited data, making them prone to overfitting and lacking robustness. Recent developments in foundation models, such as SAM-Med3D, offer a new solution by enabling powerful, generalizable feature extraction across diverse medical imaging tasks. By leveraging pretrained knowledge from large-scale medical datasets, foundation models can provide richer and more transferable representations.

\section{Methodology}
\label{sec:method}

\begin{figure}[!htb]
    \centering
    \includegraphics[trim=40 10 210 10,clip,width=\linewidth]{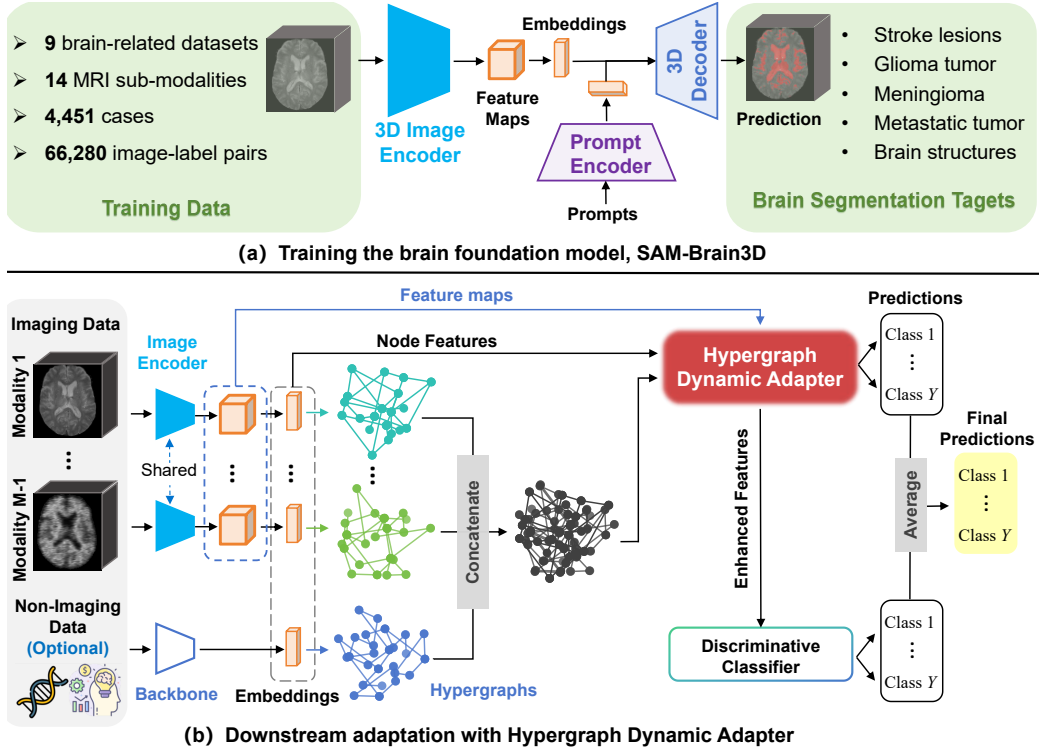}
    \caption{The pipeline of our method. (a) Training the brain foundation model of SAM-Brain3D. SAM-Brain3D is trained on diverse brain datasets and numerous image-label pairs for flexible segmentation targets. It shares the same network architecture as SAM-Med3D \cite{wang2023sam} which has an image encoder, a prompt encoder, and a decoder. (b) Downstream adaptation with Hypergraph Dynamic Adapter (HyDA) for brain disease diagnosis. This stage uses the parameter-fixed image encoder of SAM-Brain3D to extract feature maps/embeddings from multi-modal imaging data while for non-imaging data (optional), a simple Multi-Layer Perceptron (MLP)  can be used as a backbone for feature embedding. The embeddings are  exploited to construct modality-specific sub-hypergraphs, concatenated as the final hypergraph. The feature maps, the embeddings (node features), and the final hypergraph are then input to HyDA to obtain predictions and enhanced features, which are fed into a discriminative classifier to predict the disease types. We further average both predictions as the final one.
    }
    \label{fig:pipeline}
\end{figure}

As shown in Figure~\ref{fig:pipeline}, our method has two stages. 
The first stage trains SAM-Brain3D on 66,280 image-label pairs across 9 brain-related segmentation datasets. The second stage adapts the foundation model to downstream tasks, e.g., brain disease classification, using a novel Hypergraph Dynamic Adapter (HyDA).

\subsection{SAM-Brain3D: A Brain Foundation Model}
As depicted in Figure~\ref{fig:pipeline}(a), the brain foundation model, termed SAM-Brain3D, is trained on 9 brain-related segmentation datasets, encompassing 14 MRI sub-modalities and 66,280 image-label pairs from 4,451 cases. It can segment diverse targets to facilitate brain disease analysis, including stroke lesions, glioma tumors, meningioma, metastatic tumors, and 35 brain structures.  The data details are summarized in Table~\ref{tab:brain_data}.  Owing to diverse datasets and massive training data, SAM-Brain3D can excel in segmenting both seen and unseen targets, achieving promising performance and manifesting strong generalization ability (see Table~\ref{tab:cmp_iseg19}). SAM-Brain3D can be directly used for brain segmentation tasks without further adaptation. 

\begin{table}[htb]
    \centering
    \caption{Data specifications for the training of SAM-Brain3D. \#Mod. denotes the number of MRI sub-modalities and \#Pairs denotes the number of image-label pairs. \textsuperscript{1} The official dataset comprises 1,251 cases with labels but 90 cases are randomly chosen for testing.  \textsuperscript{2} The total number of sub-modalities cannot be obtained by simply summing the sub-modalities in each dataset as they may overlap.}
    \resizebox{\textwidth}{!}{
    \begin{tabular}{c|p{0.65cm}p{0.7cm}p{0.9cm}|l}
    \toprule
    \textbf{Datasets} & \#Mod. & \#Cases & \#Pairs & \textbf{Segmentation Targets}\\
    \midrule
        Learn2Reg2022 OASIS~\cite{hugo2017longitudinal} & 1 & 414 & 14,486 & 35 Brain Structures\\
        BraTS2023-MEN & 4 & 1,000 & 8,156 & 3 Meningioma Tumors\\
        BraTS2023-MET & 4 & 238 & 2,528 & 3 Metastatic Tumors\\
        BraTS2023-GLI & 4 & 1,161\textsuperscript{1} & 14,704 & 3 Adult Glioma Tumors\\
        BraTS2023-SSA & 4 & 43 & 504 & 3 African Adult Glioma Tumors\\
        BraTS2023-PED & 4 & 99 & 976 & 3 Pediatric Glioma Tumors\\
        UCSF-PDGM~\cite{calabrese2022university} & 12 & 501 & 23,777 & 4 Diffuse Glioma Tumors\\
        ATLASR2~\cite{liew2022large} & 1 & 655 &655 & Stroke Lesion\\
        ISLES2022 & 2 & 250 & 494 & Stroke Lesion\\
    \midrule
        Total & 14\textsuperscript{2} & 4,451 & 66,280 & -\\
    \bottomrule
    \end{tabular}
    }
    \label{tab:brain_data}
\end{table}

We build our SAM-Brain3D upon SAM-Med3D \cite{wang2023sam}. Among the existing foundation models for medical segmentation \cite{ma2024segment,cheng2023sam}, SAM-Med3D is chosen for the following reasons. 1) Since brain imaging data, crucial for brain disease analysis, are usually three-dimensional and multi-modal, SAM-Med3D, a thorough 3D structure trained on massive 3D multi-modal data, can better process these data. 
2) As a segmentation foundation model, SAM-Med3D can capture low-level and high-level features from medical images, both of which are important for many downstream tasks like Alzheimer's disease (AD) diagnosis. For instance, both the low-level features (e.g., volumetric changes of gray matter in the hippocampus) and high-level ones, e.g., semantic disease grading information, are necessary for correctly diagnosing AD. 
3) SAM-Med3D achieves impressive performance with strong generalization ability. State-of-the-art segmentation performance can underpin better brain structure segmentation while the promising generalizability contributes better to diverse downstream tasks, like brain-related disease diagnosis. Both properties can foster a stronger brain foundation model.

Our SAM-Brain3D shares the same architecture as SAM-Med3D \cite{wang2023sam}, consisting of an image encoder, a prompt encoder, and a decoder. The image encoder extracts feature embeddings from input images. 
The prompt encoder encodes prompts as embeddings which are provided by users, such as points and masks, to help locate the region of interest. The decoder takes both feature and prompt embeddings to predict the segmentation mask.

To tailor SAM-Brain3D for brain-specific tasks, we first initialize it with the weights from SAM-Med3D-Turbo and train all its parameters using 9 brain-related datasets in Table~\ref{tab:brain_data}. This approach allowed us to leverage the general medical imaging knowledge from SAM-Med3D while adapting the model to brain-specific segmentation tasks. Details can be found in Section~\ref{sec:exp_setup_sambrain3d}.

\subsection{Hypergraph Dynamic Adapter for Downstream Tasks}
\begin{figure}[!htb]
    \centering
    \includegraphics[trim=5 130 275 0,clip,width=\linewidth]{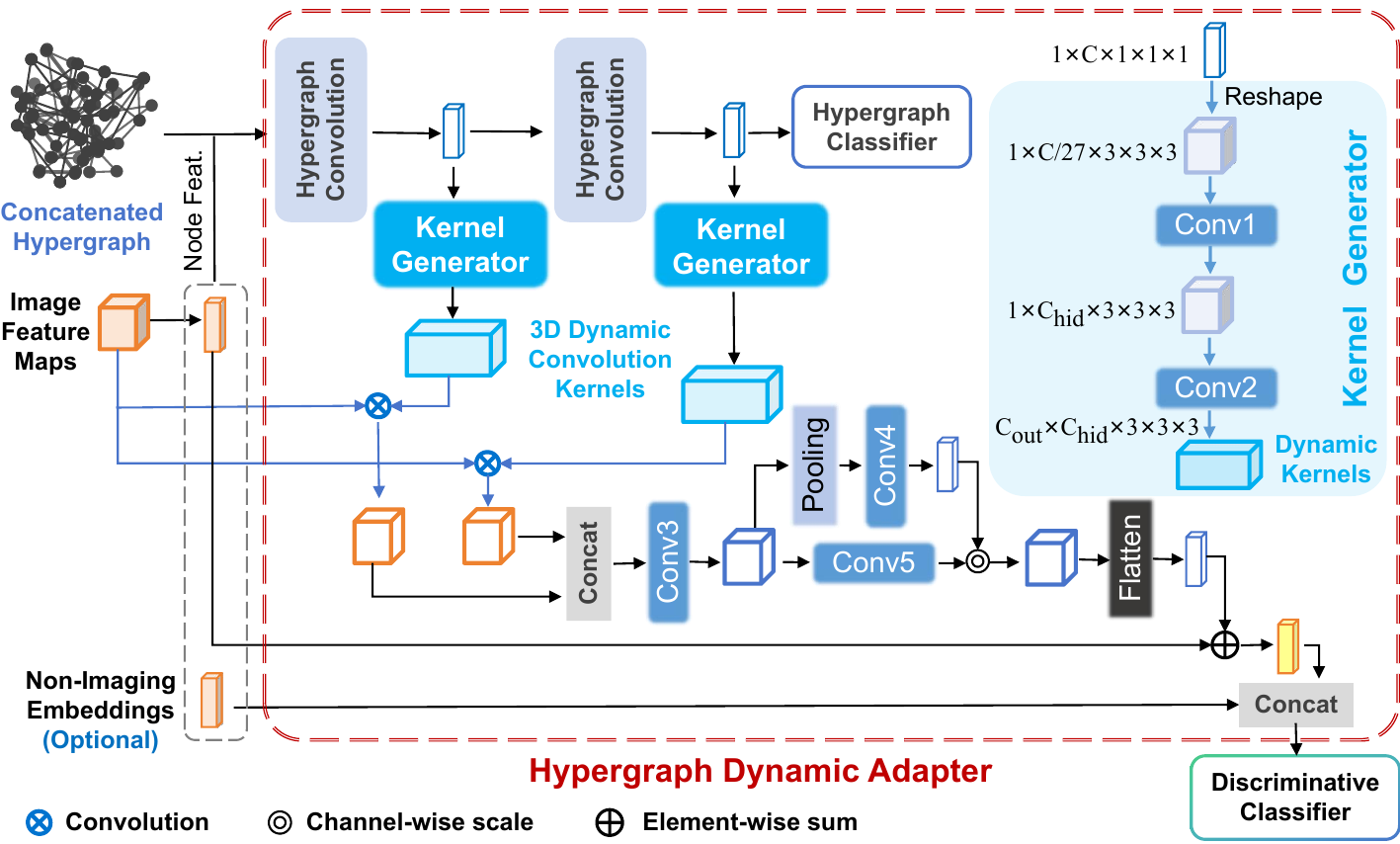}
    \caption{The design of Hypergraph Dynamic Adapter (HyDA) for downstream brain disease diagnosis tasks. HyDA exploits two hypergraph convolution layers to extract high-order relations from the final hypergraph and multi-modal feature embeddings (i.e., node features). It then makes a prediction using a hypergraph classifier. Since the semantic prediction of disease types is based on embeddings, they usually contain semantic global context. The semantic embeddings are input to kernel generators, comprising two 1$\times$1$\times$1 convolution layers (`Conv1' and `Conv2'), to generate semantic and dynamic convolutional kernels for each input subject. The subject-conditioned dynamic kernels then convolve the feature maps of imaging data to fuse the semantic information into the lower-level features, leading to dynamic multi-scale fusion. The fused feature maps are concatenated together and merged using another 1$\times$1$\times$1 convolution (`Conv3'). The merged version will be further enhanced by channel-wise reduction and scale via `Conv4\&5'. The enhanced feature maps are then flattened into a vector and summed with the original node features via a residual connection as the final feature embeddings. The final embeddings of different modalities are then concatenated and input to a discriminative classifier for prediction.
    }
    \label{fig:hyda}
\end{figure}

We propose a novel Hypergraph Dynamic Adapter (HyDA) to adapt SAM-Brain3D to downstream tasks. Since most downstream tasks are about brain disease diagnosis (with or without a specific brain disease), we will focus on classifying brain diseases. 
Figure~\ref{fig:pipeline}(b) depicts the downstream adaptation with HyDA. We use the image encoder of SAM-Brain3D for the feature extraction of imaging modalities and fix its parameters for efficient adaptation. Our framework can also take non-imaging data (optional) as input, such as age, gender, and cognitive scores. The feature embeddings of non-imaging data can be extracted leveraging a simple MLP. The embeddings of both non-imaging and imaging data are utilized to construct modality-specific sub-hypergraphs. These sub-hypergraphs are then fused or concatenated as the final hypergraph. After that, the final hypergraph, the feature maps of imaging modalities, and the feature embeddings of all the modalities will be input to HyDA to obtain the predictions and the enhanced final features. The final features are also fed into a discriminative classifier (e.g., a fully connected layer followed by SoftMax activation function) to predict the disease types. We further average both predictions to make the final diagnosis.

\textbf{Hypergraph Dynamic Adapter} (HyDA) is a lightweight plug-in module designed to have three key features: 1) \textbf{Multi-modal} relation extraction. The HyDA exploits hypergraphs to model high-order relations between different modalities so that complementary information can be extracted for better feature representations. 2) \textbf{Multi-scale} feature fusion. Higher-level information, e.g., semantic global context, will be fused into lower-level features, e.g., texture and shape, as they can be essential for better performance on downstream tasks. 3) \textbf{Dynamic} parameters for subject-wise adaptation. Since the subjects may vary in demographics (e.g., age, gender, and race) and the device for capturing their imaging data may differ in protocols and settings, dynamic parameters, inspired by \cite{deng2022dynamic}, conditioned on each subject can tackle subject-wise differences and bring better adaptation results. Next, we will elaborate on HyDA.

\paragraph{Hypergraphs for Multi-Modal Data} HyDA leverages hypergraphs to model multi-modal data rather than graphs because hypergraphs alleviate the limitation of graphs (i.e., modeling only pair-wise relation) and excel in modeling high-order relations in multi-modal data. HyDA first constructs hypergraphs using the feature embeddings of different modalities $X_1, ..., X_M$, with $M$ denoting the total modality number. A modality-specific hypergraph $G_m$ is then constructed using $X_m=\{x_m^n\}_{n=1}^N$, where $x_m^n$ denotes the features of the $n$-th subject's $m$-th modality and $N$ is the total subject number. $G_m$ has a vertices set $V_m$, a hyperedge set $E_m$, and an adjacent matrix $H_m$, namely $G_m=(V_m, E_m, H_m)$.  Each vertex $v_m^n$ in $V_m$ denotes a subject whose features are $x_m^n$. Each hyperedge captures the relations between different vertices/subjects. Unlike an edge in graphs connecting only two vertices, a hyperedge can join more than two vertices, thus capturing the high-order relations among vertices, e.g., their shared properties of brain diseases. In our implementation, each hyperedge $e_m^n$ connects $k$ nearest neighbors center around a given vertex $v_m^n$, i.e., $e_m^n=\{NN_k(v_m^n)|v_m^n\in V_m\}$. The hyperedges are concatenated together to obtain the hyperedge set for each modality $E_m$. The adjacent matrix $H_m \in \mathbb{R}^{|V_m| \times |E_m|}$ defines the weight of each vertex and hyperedge as
$h_m(v_m,e_m)= \left\{ 
    \begin{array}{lc}
        1, \ &\text{if} \  v_m \in e_m \\
        0, \ & \text{otherwise}
    \end{array}
\right.
$.

After constructing the modality-specific hypergraphs, we concatenate them as the final hypergraph $G=\text{Concat}(G_1, ..., G_M)$ to facilitate the extraction of multi-modal information, as shown in Figure~\ref{fig:hyda}. Concretely, we input the final hypergraph and the features of vertices, $(G, (X_1, ..., X_M))$, into spatial hypergraph convolution layers which are presented in \cite{gao2022hgnn} and formulated as
\begin{equation}
\label{eq:hg_conv}
    f=\text{HGConv}(G, (X_1, ..., X_M)),
\end{equation}
where $f$ is the output features of the hypergraph convolution $\text{HGConv}$. The hypergraph convolution aggregates messages via a two-step pass: passing from vertex to hyperedge and from hyperedge to vertex. This two-step strategy  \cite{gao2022hgnn} can effectively capture the high-order relations across vertices/subjects, so complementary multi-modal information is extracted to comprehensively reflect the brain functions, further producing better predictions of $p_g$. $p_g$ is obtain using $\text{HGConv}$ and a SoftMax activation function $\phi$. 
\begin{equation}
\label{eq:hg_cls}
    p_g=\phi\left(\text{HGConv}(G, f)\right).
\end{equation}

\paragraph{Multi-scale and Dynamic Feature Fusion} As shown in Figure~\ref{fig:hyda}, we will use the feature embeddings from two hypergraph convolution layers, obtained as per Equation~\eqref{eq:hg_conv}, to enforce multi-scale and dynamic fusion. Let us denote feature embeddings of the $n$-th subject from the first convolution as $f_1^n$ and these from the second as $f_2^n$. Since these two features are high-level features exploited for brain disease prediction, they usually capture semantic global context information. We then integrate their semantic information into lower-level feature maps, which mainly contain morphological or texture information. This multi-scale fusion is achieved by dynamic convolution kernels generated from $f_1^n, f_2^n$, detailed below.

We design kernel generators to generate dynamic kernels from $f_1^n, f_2^n$. Here, `dynamic' means that the parameters of these kernels are not fixed during inference and they are dynamically generated for each specific subject rather than being shared across all the subjects. The dynamic design can help HyDA tackle subject-wise variations and better adapt to each subject for precise and personalized subject-specific prediction. We will take $f_1^n$ as an example to illustrate the generation of dynamic kernels, but similar operations can be applied to $f_2^n$. Concretely, for a batch of feature embeddings $f_1\in\mathbb{R}^{B\times C\times 1 \times 1\times 1}$ ($B$ as the batch size, $C$ as the channels, and $1 \times 1\times 1$ denoting the embeddings from a 3D image), the kernel generator first reshapes the feature embeddings of the $n$-th subject $f_1^n\in\mathbb{R}^{1\times C\times 1 \times 1\times 1}$ into $\hat{f}_1^n\in\mathbb{R}^{1\times C/27\times 3 \times 3\times 3}$. 
We set the output feature dimension of hypergraphs $C$ to 864 divisible by 27. 
Then, we increase its channel to $C_{hid}$ using a 1$\times$1$\times$1 convolution layer, i.e., `Conv1' in the top right panel of Figure~\ref{fig:hyda}, leading to the shape of the output feature map being $1\times C_{hid}\times 3 \times 3\times 3$ where $C_{hid}$ is the hidden channel dimension. We then swap its first two dimensions, leading to $C_{hid}\times 1\times 3 \times 3\times 3$, and apply another convolution `Conv2' to increase its channel to match the desired output channel $C_{out}$ ($C_{out}$=128 in our case). The output feature maps becomes the shape of $C_{hid}\times C_{out}\times 3 \times 3\times 3$. Finally, we swap the first two dimensions back to obtain the 3D dynamic kernels $W_1^n \in \mathbb{R}^{C_{out}\times C_{hid}\times 3 \times 3\times 3}$. Formally, the kernel generator is formulated as
\begin{equation}
    W_1^n=g_\theta(\hat{f}_1^n),
\end{equation}
where $g_\theta$ denotes the function of kernel generator and $\theta$ is the parameter set of these two 1$\times$1$\times$1 convolution layers (namely `Conv1' and `Conv2'). 

The convolutional kernels $W_1^n$ have the following advantages. Firstly, $W_1^n$ are \textbf{dynamic kernels conditioned on each subject} $n$ as different subjects have distinct feature $\hat{f}_1^n$, leading to non-identical kernels of $W_1^n$ for different subject $n$. Owing to $W_1^n$, HyDA can address subject-wise variations and better adapt to each subject for precise and personalized subject-specific prediction.
Secondly, $W_1^n$ contains semantic information inherited from $\hat{f}_1^n$, enabling \textbf{better multi-scale fusion via convolving with low-level feature maps}:
\begin{equation}
\label{eq:dynamic_conv}
    O_{1,m}^n = \sigma(W_1^n \otimes F_{m}^n) = \sigma\left (g_{\theta}(\hat{f}_1^n)\otimes F_{m}^n \right ),
\end{equation}
where $O_{1,m}^n$ denotes the output feature map and $\sigma$ represents the ReLU activation function. $\otimes$ is the convolution operation while $F_{m}^n\in\mathbb{R}^{1\times C_{hid}\times D\times H \times W}$ is the feature map of the $n$-th subject obtained from the image encoder of SAM-Brain3D, with $D,H,W$ denoting depth, height, and width. Equation~\eqref{eq:dynamic_conv} enables $W_1^n$ to interact and transform low-level features in a non-linear and subject-adaptive manner, thus more effective in multi-scale fusion compared with simply averaging or concatenating multi-scale features.  
Since $W_1^n$ contains the semantic global context inherited from $\hat{f}_1^n$ (or $f_1^n$ as the reshape operation does not change the semantic information), the dynamic convolution in Equation~\eqref{eq:dynamic_conv} can integrate the semantic global context of $W_1^n$ into morphological information of $F_{m}^n$ for better brain disease diagnosis. 

After obtaining the dynamically generated feature maps $O_{1,m}^n, O_{1,m}^n$ from $f_1^n, f_2^n$, respectively, we concatenate and merge them via a convolution layer `Conv3', which results in $O_{m}^n\in\mathbb{R}^{1\times C_{hid}\times D\times H \times W}$ sharing the same shape as $F_{m}^n$. $O_{m}^n$ is then enhanced by channel-wise reduction and scale via `Conv4' and `Conv5', similar to the Squeeze-and-Excitation (SE) block~\cite{hu2018squeeze}. The channel reduction ensures that the flattened feature map can match the shape of the original input feature embeddings $x_m^n\in X_m$ to facilitate the element-wise sum. These steps can be formulated as
\begin{equation}
\begin{aligned}
    O_{m}^n&=g_{\theta_3} \left (\text{Concat}(O_{1,m}^n, O_{1,m}^n) \right ),\\
    \hat{O}_{m}^n&=\psi\left (g_{\theta_4}(\text{Pool}(O_{m}^n)) \right)  \odot \sigma \left (g_{\theta_5}(O_{m}^n)\right ),\\
    \tilde{f}_m^n &= x_m^n+\text{Flatten}(\hat{O}_{m}^n),
\end{aligned}
\end{equation}
where $g_{\theta_3}, g_{\theta_4}, g_{\theta_5}$ denote the `Conv3'$\sim$`Conv5', with $\theta_3,\theta_4,\theta_5$ as their learnable parameters. $\psi$ stands for the sigmoid activation function while $\sigma$ for ReLU. $\odot$ represents channel-wise scale and $\text{Flatten}$ is the operation of reshaping input into a vector. $\tilde{f}_m^n$ is the enhanced feature representation to be used for a discriminative classifier for predicting brain diseases.

\textbf{Remark.} Though the kernels $W_1^n$ are conditioned on each subject $n$, we do not learn a separate set of parameters for every subject. Instead, we design a lightweight \textit{kernel generator} composed of only two standard 1$\times$1$\times$1 convolution layers (denoted as `Conv1’ and `Conv2’ in Figure~\ref{fig:hyda}) to generate these subject-specific kernels. This generator has only $C/27*C_{hid}+C_{out}$ parameters, making it highly efficient. For instance, in our implementation, $C$=864, $C_{hid}$=384, and $C_{out}$=128, so the total parameters are 864$\times$384$+$128$\approx$0.3M. 
From another perspective, the generated kernels $W_1^n$ can be interpreted as special feature maps output by the kernel generator, dynamically conditioned on each subject. Since kernel generation is essentially a feature map transformation using conventional convolutions, we observe no stability issues during training.

\subsection{Optimization Scheme}
We have two predictions, $p_g$ from the hypergraph classifier and $p_d$ from the discriminative classifier (comprising a fully connected layer followed by a SoftMax activation function). $p_g$ and  $p_d$ are utilized to learn HyDA by optimizing the objective as
\begin{equation}
    L=\frac{1}{N}\sum_{n=1}^N \left [\text{CE}(p_g^n, y^n) + \text{FL}(p_g^n, y^n) + \text{CE}(p_d^n, y^n) + \text{FL}(p_d^n, y^n)\right ],
\end{equation}
where $L$ is the total loss and $CE$ is the cross-entropy loss. $FL$ is focal loss~\cite{lin2017focal} to tackle the class imbalance problem. Then, the average prediction $p=(p_g+p_d)/2$ is used as the final prediction. 
Here, we treat the predictions of hypergraph and discriminative classifiers equally because \#1 and \#2 of Table~\ref{tab:ablation} show that they generally achieve similar performance, particularly in accuracy and Specificity. Thus, it is reasonable to treat them equally can avoid tuning hyperparameter to control their weight.

\section{Dataset and Experimental Setup}
In this section, we introduce the datasets and experimental setups of training the brain foundation model, SAM-Brain3D, as well as fine-tuning the Hypergraph Dynamic Adapter (HyDA) for two downstream tasks, including Alzheimer's progression and O6-methylguanine-DNA methyltransferase (MGMT) classification.

\subsection{Experimental Setup of SAM-Brain3D}
\label{sec:exp_setup_sambrain3d}

\paragraph{Dataset Description}
To integrate specialized brain region knowledge into the foundation model, we first initialize SAM-Brain3D using the weights from SAM-Med3D-Turbo, which is trained on both brain and non-brain images. We then train all its parameters on 9 brain-specific datasets (detailed in Table~\ref{tab:brain_data}).  These datasets encompass 14 distinct MRI modalities and over 66,000 image-label pairs of 4,451 cases, enabling the model to capture detailed anatomical and modality-specific characteristics of the brain. Specifically, we exploit the following datasets. 
1) \textit{Learn2Reg2022 OASIS}: 
The OASIS dataset~\cite{hugo2017longitudinal} from the Learn2Reg Challenge\footnote{https://learn2reg.grand-challenge.org/Datasets/} contains 35 different brain structures, including ventricles, vessels, and cerebellar white matter. It provides T1-weighted MRI scans from 455 cases.
2) \textit{BraTS2023-MEN}: 
The BraTS2023 Meningioma Challenge dataset\footnote{https://www.synapse.org/Synapse:syn51156910/wiki/622353} aims to segment meningiomas from multiparametric MRI. It comprises 1,650 cases from six centers, with 1,000 cases and their corresponding meningioma segmentation results for training. It provides four MRI sub-modalities, including pre- and post-gadolinium T1-weighted (labeled as T1 and T1CE), T2-weighted (T2), and T2-weighted fluid attenuated inversion recovery (T2-FLAIR), with these four shared by the other BraTS2023 datasets presented below.
3) \textit{BraTS2023-MET}: 
The BraTS2023 Brain Metastases Challenge dataset\footnote{https://www.synapse.org/Synapse:syn51156910/wiki/622553} focuses on segmenting brain metastases from multiparametric MRI. It contains 328 cases, with 238 annotated cases in the training set, each including four MRI sequences and corresponding brain metastases segmentation results. 
4) \textit{BraTS2023-GLI}: 
The BraTS2023-GLI (BraTS21) dataset\footnote{https://www.synapse.org/Synapse:syn51156910/wiki/622351}~\cite{bakas2017advancing,menze2014multimodal} is a large-scale multi-modal MRI dataset for brain glioma segmentation with over 1,000 cases. All the data and labels are preprocessed, aligned with a unified anatomical template, and adjusted to 1 mm$^3$ spacing. 
5) \textit{BraTS2023-SSA}: 
The BraTS2023 Sub-Saharan Africa Challenge dataset\footnote{https://www.synapse.org/Synapse:syn51156910/wiki/622556} focuses on glioma segmentation in patients from the Sub-Saharan African region. It includes 43 annotated training samples with glioma segmentation results. 
6) \textit{BraTS2023-PED}: The BraTS2023 Pediatrics Tumor Challenge dataset\footnote{https://www.synapse.org/Synapse:syn51156910/wiki/622461} aims to segment pediatric tumors. It collects 228 pediatric high-grade gliomas from multiple institutions, with 99 for training. 
7) \textit{UCSF-PDGM}: 
The UCSF Preoperative Diffuse Glioma MRI dataset\footnote{https://www.cancerimagingarchive.net/collection/ucsf-pdgm/} ~\cite{calabrese2022university} comprises 501 cases (including some cases overlapped with \textit{BraTS2023 GLI}) for diffuse glioma segmentation. Each case features 12 different MRI modalities and provides both tumor and parenchyma segmentation. 
8) \textit{ATLASR2}: 
The ATLASR2 dataset~\cite{liew2022large} is a comprehensive collection for stroke lesion segmentation from single-modality T1-weighted MRI. It offers 655 public images with annotation. 
9) \textit{ISLES22}: 
The ISLES22 dataset\footnote{https://isles22.grand-challenge.org/} focuses on the automatic segmentation of acute to subacute ischemic stroke lesions using FLAIR, DWI, and ADC MRI modalities. It collects 400 cases from multiple centers using different devices, divided into 250 training cases and 150 test cases.

We randomly select 1,161 patient in \textit{BraTS2023 GLI} for training SAM-Brain3D. The remaining 90 annotated cases are used for testing our SAM-Brain3D and SAM-Med3D. We also compare SAM-Brain3D with SAM-Med3D on unseen categories of iSeg19 dataset~\cite{sun2021multi} to evaluate their generalization ability for brain segmentation tasks. iSeg19 dataset seeks to segment infant brain MRI into white matter (WM), gray matter (GM), and cerebrospinal fluid (CSF).

\paragraph{Implementation Details}
We set the batch size to 12 and the initial learning rate to $8 \times 10^{-5}$, which is 10\% of the learning rate used for SAM-Med3D training. The model is trained for 200 epochs using Adam optimizer. We further reduce the learning rate by 10 at the 120th and 180th epochs.

\subsection{Experimental Setup of Alzheimer's Progression}
Alzheimer's disease (AD) is a progressive brain disease that causes memory loss, cognitive decline, and even death. The symptoms are usually mild at the early stage, e.g., Mild Cognitive Impairment (MCI) stage, and gradually become more severe over several years. Since MCI patients may convert to AD  (progressive MCI) or remain stable (stable MCI), it is important to distinguish these two categories for precise medical intervention.  The task of classifying progressive MCI (pMCI) and stable MCI (sMCI) is called the Alzheimer's progression task. Though AD is not curable, correctly identifying pMCI patients enables early treatment to slow down the progression of converting to AD, thus improving patient outcomes.

\paragraph{Dataset Description} We use the popular Alzheimer's Disease Neuroimaging Initiative (ADNI) dataset\footnote{Data used in the preparation of this article were obtained from the Alzheimer's Disease Neuroimaging Initiative (ADNI) dataset (\url{adni.loni.usc.edu}).}~\cite{jack2008alzheimer} for Alzheimer's progression. We first identify the subjects with MCI at the baseline visit, and seek to predict whether MCI subjects will convert to AD (i.e., pMCI) or keep stable (i.e., sMCI) in a two-year window. We utilize both imaging and non-imaging modalities for this task. The imaging modalities include MRI and PET scans while the non-imaging one contains age, gender, education year, the genetic variant of APOE4, and cognitive scores of ADNI-MEM \cite{crane2012development} and ADNI-EF \cite{gibbons2012composite}. After filtering, we have 190 MCI subjects with complete three modalities ($M$=3) and labels (i.e., diagnosis results in the two-year window). Among these subjects, 146 are sMCI and 44 are pMCI. We then split the subjects into training and validation sets with a ratio of 8:2 and conduct 5-fold cross-validation to evaluate our method.

\paragraph{Implementation Details} We pre-process the 3D MRI volumes as follows: 1) aligning anterior commissure with posterior commissure, 2) removing skull, 3) correcting intensity, 4) removing cerebellum, and 5) applying linear alignment to a template MRI. The PET volumes are aligned with corresponding MRI volumes via linear registration. The MRI and PET volumes are center-cropped into 128$\times$128$\times$128 and randomly flipped for data augmentation. Then the images and each feature in non-imaging data are normalized to [0,1].

We fix the image encoder from SAM-Brain3D and train HyDA using AdamW optimizer, with an initial learning rate of 0.001 and batch size of 30. The hypergraphs for HyDA are constructed using $k=20$ nearest neighbors by default and $k$ will be evaluated in our experiments. 
We randomly  drop out 50\% vertex features of hypergraphs as a regularization. When optimizing hypergraph convolution layers, we set the weight decay of the AdamW optimizer to 0.01 as another regularization. 
We run all the experiments on a NVIDIA A100 GPU with 80 GB memory using PyTorch~\cite{paszke2019pytorch}.

\subsection{Experimental Setup of MGMT Classification}
O$^6$-methylguanine-DNA methyltransferase (MGMT) promoter methylation is a key biomarker in glioblastoma, strongly associated with better prognosis and enhanced response to alkylating chemotherapy \cite{lechapt2012o6}. Accurate prediction of MGMT status can guide personalized treatment decisions and reduce the need for invasive surgical biopsies. Developing non-invasive imaging-based methods, such as radiogenomic models using MRI, has the potential to accelerate diagnosis, optimize therapy, and improve clinical outcomes for glioblastoma patients \cite{baid2021rsna}.

\paragraph{Dataset Description}
We evaluate the proposed method on the Brain Tumor Radiogenomic Classification Challenge dataset, aiming to predict MGMT promoter methylation status from glioblastoma MRI scans \cite{baid2021rsna}. The dataset contains 585 de-identified cases collected from public and institutional sources, including TCGA-GBM, ACRIN-FMISO-Brain (ACRIN 6684), and the TCIA archive \cite{clark2013cancer}. Each case includes four multiparametric MRI (mpMRI) modalities: T1, Gadolinium-enhanced T1wCE, T2, and T2-FLAIR, providing complementary information for tumor localization. To ensure consistency in slice number and benefit from standardized pre-processing, we replace the original DICOM-format data with matched NIfTI-format images from the BraTS segmentation dataset. This substitution ensures uniform volume dimensions and continuous slices across modalities. Patients with IDs [00109], [00123], and [00709] are excluded due to poor image quality, resulting in 574 cases for experiments.

\paragraph{Experimental Settings}We adopt a 5-fold cross-validation strategy, with 459 patients for training and 115 for validation in each fold. All models are trained for 25 epochs using the Adam optimizer with an initial learning rate of 0.0001. A learning rate warm-up is applied during the first 3 epochs, followed by decay at epoch 6 with a gamma of 0.5. Early stopping with a patience of 7 epochs is employed. Input volumes are resized to $128 \times 128 \times 128$. We use random flipping for data augmentation. 

We compare CNN-based models (EfficientNet-B3-3D \cite{tan2021efficientnetv2,saeed2023mgmt}, ResNet50-3D \cite{chen2019med3d,saeed2023mgmt}, and the encoder of SegResNet \cite{myronenko20183d}), a transformer-based model (Swin-ViT3D \cite{he2023swinunetr,saeed2023mgmt}), graph-based models (ViG-3D \cite{hu2024mgmt} and Population-Graph \cite{zhang2023population}), and hypergraph-based models. CNN-based and Swin-ViT3D models are trained with a batch size of 8, while graph and hypergraph models are trained with a batch size of 20.

For graph construction, both basic graphs and population graphs are built using $k=8$ nearest neighbors, while hypergraphs are constructed with $k=16$ nearest neighbors. In CNN-based and Swin-ViT3D models, two fully connected layers with ReLU activation are appended after the encoder. For graph and hypergraph models, two GCNConv layers with ReLU are applied after feature extraction. EfficientNet-B3-3D is trained from scratch due to the lack of publicly available 3D pretrained weights. ResNet50-3D uses Med3D pretrained weights, which is trained on multiple MRI and CT tasks \cite{chen2019med3d}. The encoders of SegResNet and Swin-ViT3D are initialized with pretrained weights from \cite{myronenko20183d} and \cite{zhang2024biophysics}, respectively. All graph and hypergraph models employ the frozen SAM-Brain3D encoder as a feature extractor.

\section{Results}
This section presents the experimental results of SAM-Brain3D for brain segmentation tasks and HyDA for two downstream brain disease tasks: Alzheimer's progression and MGMT classification.

\begin{table}[htb]
    \centering
    \caption{Comparison with state-of-the-art foundation models on BraTS21 dataset. NE. Tumor: Non-enhancing Tumor, E. Tumor: Enhancing Tumor. The best performance of each modality is in bold.}
    \begin{tabular}{c|c|ccc|c}
    \toprule
      Modalities   & Methods & Edema	&NE. Tumor	&E. Tumor
 & Avg. \\
    \midrule
       \multirow{2}{*}{MRI-T1}  & SAM-Med3D Turbo & 41.42	&15.87	&45.09	&34.13\\
        & SAM-Brain3D & \textbf{50.05}	&\textbf{17.36}	&\textbf{49.31}	&\textbf{38.91}\\
    \midrule
        \multirow{2}{*}{MRI-T1ce}  & SAM-Med3D Turbo & 49.39	&26.82	&59.31	&45.18 \\
        & SAM-Brain3D & \textbf{53.24}	&\textbf{29.04}	&\textbf{61.37}	&\textbf{47.88}\\
    \midrule
       \multirow{2}{*}{MRI-T2}  & SAM-Med3D Turbo & 53.75	&21.34	&49.86	&41.65\\
        & SAM-Brain3D & \textbf{58.95}	&\textbf{24.26}	&\textbf{50.08}	&\textbf{44.43}\\
    \midrule
        \multirow{2}{*}{MRI-FLAIR}  & SAM-Med3D Turbo & \textbf{65.45}	&\textbf{22.70}	&\textbf{58.00}	&\textbf{48.72} \\
        & SAM-Brain3D & 61.93	&19.42	&50.98	&44.11\\
    \midrule
        \multirow{2}{*}{\textbf{Average}}  & SAM-Med3D Turbo & 52.50	& 21.69	& \textbf{53.07}	& 42.42
 \\
        & SAM-Brain3D & \textbf{56.04}	& \textbf{22.52}	& 52.94	& \textbf{43.83}\\
    \bottomrule
    \end{tabular}
    \label{tab:cmp_brats21}
\end{table}

\subsection{SAM-Brain3D for Brain Segmentation Tasks}

We compare SAM-Brain3D with SAM-Med3D on both seen and unseen categories, corresponding to BraTS21 and iSeg19~\cite{sun2021multi}, respectively. We adopt the Dice score (\%) to evaluate the segmentation performance.

Table~\ref{tab:cmp_brats21} compares SAM-Brain3D with SAM-Med3D on BraTS21 where all its categories are seen during training. We observe a clear better Dice score of SAM-Brain3D over SAM-Med3D-Turbo on T1, T1ce, and T2 modalities for all three categories. For instance, SAM-Brain3D surpasses SAM-Med3D by 8.63\% when segmenting Edema using T1 MRI, and shows 4.78\% improvement in terms of averaged Dice for T1 modality. SAM-Brain3D is inferior to SAM-Med3D on the FLAIR modality, probably because FLAIR is less represented in our training set. Despite this, our SAM-Brain3D achieves better average Dice for all four modalities (the last two rows), suggesting its strong potential in brain disease analysis.

\begin{table}[htb]
    \centering
    \caption{Comparison with state-of-the-art foundation models on iSeg19 dataset~\cite{sun2021multi}. Dice score (\%) is reported. CSF: Cerebrospinal Fluid, GM: Gray Matter, WM: White Matter.}
    \begin{tabular}{c|c|ccc|c}
    \toprule
      Modalities   & Methods & CSF	& GM	& WM & Avg. \\
    \midrule
       \multirow{2}{*}{MRI-T1}  & SAM-Med3D Turbo & 3.64	&16.85	&51.25	&23.91\\
        & SAM-Brain3D & \textbf{15.20}	&\textbf{38.47}	&\textbf{41.55}	&\textbf{31.74}\\
    \midrule
        \multirow{2}{*}{MRI-T2}  & SAM-Med3D Turbo & 9.86	&15.78	&20.12	&15.25 \\
        & SAM-Brain3D & \textbf{15.79}	&\textbf{33.76}	&\textbf{36.74}	&\textbf{28.76}\\
    \midrule
        \multirow{2}{*}{\textbf{Average}}  & SAM-Med3D Turbo & 6.75	& 16.31	& 35.68	& 19.58 \\
        & SAM-Brain3D & \textbf{15.49}	& \textbf{36.12}	& \textbf{39.15}	& \textbf{30.25}\\
    \bottomrule
    \end{tabular}
    \label{tab:cmp_iseg19}
\end{table}

Table~\ref{tab:cmp_iseg19} further presents the comparison of segmenting unseen categories from iSeg19. On all three targets and both modalities, SAM-Brain3D  demonstrates significant performance improvements, e.g., 13.51\% performance gain over SAM-Med3D for the average Dice on the T2 task. This suggests that SAM-Brain3D benefits from brain knowledge in the 9 brain-related training datasets. This prior knowledge thus effectively enhances the model's generalization ability in brain-related tasks which is crucial for diverse downstream tasks of brain disease diagnosis.

\subsection{HyDA for Alzheimer's Progression}
We will first comprehensively evaluate the key designs of HyDA and then compare HyDA with state-of-the-art methods for Alzheimer's progression. Throughout this section, averaged results (\%) over 5-fold cross-validation are reported. We report four metrics: Accuracy (ACC), F1 score, Specificity (SPE), and Sensitivity (SEN).

\paragraph{Effectiveness of Key Designs} 
We verify the effectiveness of key designs in HyDA by gradually adding each of them to a simple baseline. The baseline freezes the image encoder of SAM-Brain3D and learns a simple discriminative classifier (a fully connected layer followed by SoftMax activation function). Its results are shown in \#1 of Table~\ref{tab:ablation}. We then replace the discriminative classifier with a hypergraph-based one, defined as Equation~\eqref{eq:hg_cls}. Compared with \#1, \#2 improves F1 score by 1.16\% and Sensitivity by 1.43\%, demonstrating the superiority of hypergraph for capturing high-order relations from multi-modal data. When the predictions of these two classifiers are averaged (\#3), all four evaluation metrics are increased, particularly 2.50\% improvement of F1 and 6.32\% in terms of Sensitivity. Finally, the multi-scale and dynamic fusion of HyDA (\#4) can further boost the accuracy from 85.52\% to 88.09\%, F1 from 65.93\% 70.23\%, and Specificity from 95.17\% to 96.43\%, at the expense of slightly decreasing the Sensitivity. Overall, HyDA (\#4) manifests a clear improvement over the baseline of a simple discriminative classifier (\#1), which verifies its effectiveness.
\begin{table}[tb]
    \centering
    \caption{Verification of key designs on ADNI. Averaged results (\%) over 5-fold cross-validation are reported.}
    \begin{tabular}{c|c|cccc}
    \toprule
      \# & Methods   & ACC & F1 & SPE & SEN \\
    \midrule
      1 & Encoder + Discriminative Classifier & 85.33	&	62.27	&94.83	&55.15\\
      2 & Encoder + Hypergraph Classifier & 85.33	& 63.43	&94.83	&56.58\\
      3 & Averaged Prediction of \#2 \& \#3 &85.52 &65.93	&95.17	&\textbf{62.90} \\
      4 & \#3 + HyDA &\textbf{88.09} &	\textbf{70.23}	&\textbf{96.43}	&62.12 \\
    \bottomrule
    \end{tabular}
    \label{tab:ablation}
\end{table}

\paragraph{Superiority of SAM-Brain3D's Image Encoder} 
Though Table~\ref{tab:cmp_iseg19} and ~\ref{tab:cmp_brats21} have justified the superiority of SAM-Brain3D over SAM-Med3D for brain segmentation tasks, we further show in Table~\ref{tab:cmp_encoder} that the image encoder of SAM-Brain3D can be better transferred to downstream task for brain disease diagnosis. This is evidenced by the clear decrease in all four metrics when replacing the image encoder of SAM-Brain3D with that of SAM-Med3D. 

\begin{table}[tb]
    \centering
    \caption{Comparison of SAM-Brain3D with SAM-Med3D for downsteam adaptation on ADNI.}
    \begin{tabular}{c|cccc}
    \toprule
       Methods   & ACC & F1 & SPE & SEN \\
    \midrule
      SAM-Brain3D Encoder + HyDA &\textbf{88.09} &	\textbf{70.23}	&\textbf{96.43}	&\textbf{62.12} \\
      SAM-Med3D Encoder + HyDA &84.24	&	63.80 	&92.86	&59.57 \\
    \bottomrule
    \end{tabular}
    \label{tab:cmp_encoder}
\end{table}

\paragraph{Evaluation on the Choice of Hypergraphs} 
By default, we adopt the hypergraph and hypergraph convolution presented in Hypergraph Neural Network Plus (HGNNP) \cite{gao2022hgnn}, which is a general hypergraph framework. We evaluate this design choice by replacing HGNNP with Dynamic Hypergraph Neural Networks (DHGNN)~\cite{jiang2019dynamic} which updates hypergraph structure using both local and global features. Table~\ref{tab:cmp_choice_hg} shows that HGNNP generally achieves better performance, owing to its general design for handling complicated heterogeneous hypergraphs.

We argue that hypergraphs can better model high-order relations in multi-modal data than graphs. To support our argument, we replace hypergraphs of HyDA with graphs and use graph convolution to substitute the hypergraph convolution. We can see in Table~\ref{tab:cmp_choice_hg} that graphs fail to beat their hypergraph counterparts. This comparison underlines the superiority of hypergraphs for multi-modal modeling. We also remark that HyDA is lightweight with only 2.8M parameters, efficient for downstream adaption.

\begin{table}[tb]
    \centering
    \caption{Evaluation on the choice of hypergraphs on ADNI.}
    \begin{tabular}{c|c|cccc}
    \toprule
      Methods  &\#Params & ACC & F1 & SPE & SEN \\
    \midrule
       HyDA default (Built on HGNNP) & 2.8M &\textbf{88.09} &	\textbf{70.23}	&96.43	&\textbf{62.12} \\
       HyDA variant (HyDA-DHGNN)& 2.8M &87.83	&	68.21	&\textbf{97.14}	&59.83\\
       HyDA variant (HyDA-Graphs)& 2.8M  & 85.85	&66.07	&\textbf{97.14}	&55.63\\
    \bottomrule
    \end{tabular}
    \label{tab:cmp_choice_hg}
\end{table}

\begin{figure}[tb]
\begin{minipage}{0.55\textwidth}
\includegraphics[trim=20 50 10 5,clip,width=\textwidth]{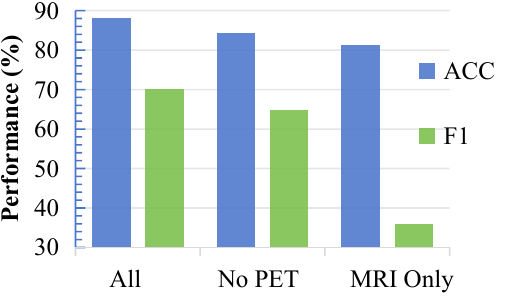}
\caption{Ablation on modalities.}
\label{fig:ablation_modality}
\end{minipage}
\begin{minipage}{0.45\textwidth}
~
\includegraphics[trim=4 5 35 5,clip,width=\textwidth]{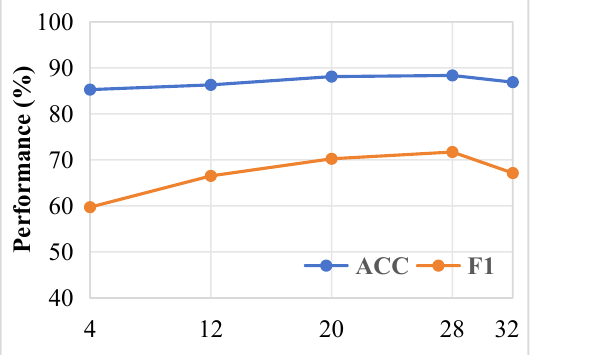}
\vspace{-0.7cm}
\caption{Evaluation on the nearest neighbor $k$.}
\label{fig:hyperparam_k}
\end{minipage}
\end{figure}

\paragraph{Ablation on Modalities}
Our HyDA effectively extracts information from three modalities, i.e., MRI, PET and non-imaging data. We then conduct the ablation on the modalities and show their results in Figure~\ref{fig:ablation_modality}. It is clear that with all three modalities, HyDA achieves the best performance, implying that HyDA can benefit from multi-modal data for better performance. Removing PET results in a clear decrease in both accuracy and F1. We attribute the performance drop to the absence of information in PET, which is essential for HyDA to fuse multi-modal information. It is also worth noting that HyDA can effectively leverage PET for better performance even though our SAM-Brain3D is trained on only MRI data. Further discarding non-imaging modalities and using MRI only can reduce both the accuracy and F1, highlighting the importance of leveraging multi-modal data for good performance.

\paragraph{Evaluation on the Number of Nearest Neighbors}
When constructing the hypergraphs, we set the number of nearest neighbors $k$ to 20 as the default setting. We further evaluate the sensitivity of the hyperparameter $K$ in Figure~\ref{fig:hyperparam_k}. We observe that the accuracy and F1 increase when $k$ varies from 4 to 28 and then decrease with a larger $k$. The optimal $k$ is 28. Overall, the accuracy is generally stable with the varying $k$ while F1 is more sensitive. Such a difference can be caused by the imbalanced class.

\paragraph{Comparison with State-of-the-Art Methods}
\begin{table}[tb]
    \centering
    \caption{Comparison with state-of-the-art methods on ADNI. \textsuperscript{1} The class balanced accuracy (BACC) is adapted from the original paper rather than ACC.}
    \begin{tabular}{c|ccccc}
    \toprule
      Methods   & ACC & F1 & SPE & SEN & AUC\\
    \midrule
      Multimodal CNN~\cite{huang2019diagnosis} &72.22 &- & 71.25 &73.44 & 77.49 \\ 
      ProAuto-noAge~\cite{shen2021heterogeneous} &78.70 & - &80.00 & \textbf{77.30} &77.56 \\
      Multi-scale graph~\cite{hett2021multi}  &80.60  &- &- &- & 85.50 \\ 
      HAN~\cite{li2023early} &71.10 & \textbf{76.66} &- &- & 76.19 \\
      VAP-Former~\cite{kang2023visual}  & 79.22\textsuperscript{1}  & 63.13 &- &- & \textbf{86.31} \\ 
      MMSDL~\cite{abdelaziz2025multi} &75.16 &61.90 & 93.81 &50.20 & 71.87 \\ 
    \midrule
       SAM-Brain3D + HyDA ($k$=20) &88.09 &	70.23	&96.43	&62.12  & 82.54\\ 
       SAM-Brain3D + HyDA ($k$=28) &\textbf{88.34} &	71.70 &\textbf{97.86}	&62.19 & 84.29\\ 
    \bottomrule
    \end{tabular}
    \label{tab:cmp_sota_ad}
\end{table}

We compare our method with the following state-of-the-art methods, all of which exploit multi-modal data. Specifically, the competitors include 1) Multimodal CNN~\cite{huang2019diagnosis} using concatenation for multi-modal fusion, 2) ProAuto-noAge~\cite{shen2021heterogeneous} utilizing auxiliary MRI and PET data for multi-modal modeling, 3) Multi-scale graph~\cite{hett2021multi} employing graphs for multi-scale fusion and a  consistent predictions across different modalities, 4) HAN~\cite{li2023early} using hypergraph for multi-modal fusion, 5) VAP-Former~\cite{kang2023visual} adopting attention-based transformers to fuse both image and attribute modalities, and 6) the latest MMSDL~\cite{abdelaziz2025multi} leveraging multi-modal attention for heterogeneous data.

Table~\ref{tab:cmp_sota_ad} shows the results of all the methods. From the results, we have the following observations. Firstly, our HyDA with the image encoder of SAM-Brain3D achieves the best accuracy and Specificity, beating the other competitors by 7.74\% in accuracy, and 4.45\% in Specificity. This large gap illustrates that our HyDA and SAM-Brain3D can outperform the other multi-modal fusion methods and obtain state-of-the-art performance for Alzheimer's progression. Secondly, compared with multi-modal fusion using simple concatenation in Multimodal CNN~\cite{huang2019diagnosis} or auxiliary data in ProAuto-noAge~\cite{shen2021heterogeneous}, our hypergraph-based dynamic fusion surpasses them in accuracy, F1 score, AUC, and Specificity, usually by a large margin. This verifies the effectiveness of our HyDA in capturing high-order relations in multi-modal data. Our method is worse than them in Sensitivity, probably due to our highly imbalanced class (146 sMCI vs. 44 pMCI). Thirdly, the accuracy of our method is superior to Multi-scale graph~\cite{hett2021multi} and the hypergraph method, HAN~\cite{li2023early}, while the F1 score is worse than HAN and the AUC is comparable to Multi-scale graph. It implies that our method can correctly classify the majority of cases, but may fall short in accurately recognizing the minority class. Finally, compared with the attention-based fusion, VAP-Former~\cite{kang2023visual} and MMSDL~\cite{abdelaziz2025multi}, our method is clearly better in accuracy and F1 score. Particularly, our method manifests a significant improvement over the latest MMSDL in all five metrics. Overall, our method is highly competitive among state-of-the-art methods.

\subsection{HyDA for MGMT Classification}

\begin{table}[ht]
\centering
\caption{Comparison of different training strategies (TFS: training from scratch, FT-FL: freezing encoder and training rest part, FT-ALL: fine-tuning all parameters) across models on 5-fold cross-validation. All results are reported as AUC scores (\%).}
\footnotesize
\resizebox{\textwidth}{!}{%
\begin{tabular}{l|c|ccccc|c}
\toprule
\multirow{2}{*}{Methods} & \multirow{2}{*}{\makecell{Strategies}} & \multicolumn{5}{c|}{Fold AUC (\%)} & Mean ± Std \\
\cmidrule(lr){3-7} \cmidrule(lr){8-8}
& & Fold 1 & Fold 2 & Fold 3 & Fold 4 & Fold 5 & AUC \\
\midrule
EfficientNet & TFS & 58.94 & 57.68 & 62.14 & 58.88 & 64.75 & 60.48±2.60 \\
\midrule
ResNet & TFS & 64.33 & 58.76 & 60.73 & \textbf{65.52} & 67.28 & 63.32±3.13 \\
& FT-FL & 57.30 & 55.76 & 56.06 & 64.42 & 55.28 & 57.76±3.39 \\
& FT-ALL & 56.70 & 52.58 & 49.30 & 60.55 & 60.56 & 55.94±4.44 \\
\midrule
SegResNet & TFS & 52.88 & 58.33 & 59.06 & 58.18 & 58.40 & 57.37±2.27 \\
& FT-FL & 54.55 & 60.73 & 51.79 & 61.55 & 56.48 & 57.02±3.69 \\
& FT-ALL & 56.39 & 62.67 & 59.61 & 60.30 & 58.83 & 59.56±2.04 \\
\midrule
Swin-ViT & TFS & 59.14 & 57.50 & 53.94 & 59.30 & 60.19 & 58.01±2.21 \\
& FT-FL & 47.15 & 58.36 & 53.67 & 58.91 & 54.78 & 54.57±4.22 \\
& FT-ALL & 59.14 & 57.50 & 53.94 & 59.30 & 60.19 & 58.01±2.21 \\
\midrule
SAM-Brain3D+GraphNet & FT-FL & 65.27 & 61.26 & 57.92 & 56.30 & 59.88 & 60.13±3.08 \\
SAM-Brain3D+PopGraphNet & FT-FL & 58.26 & 50.09 & 53.88 & 50.47 & 43.73 & 51.29±4.79 \\
SAM-Brain3D+HyDA-GCN & FT-FL & \textbf{66.45} & 60.18 & 59.70 & 61.58 & \textbf{68.12} & 63.21±3.43 \\
SAM-Brain3D+HyDA-DHGNN & FT-FL & 64.21 & 58.76 & 60.03 & 64.88 & 57.04 & 60.98±3.07 \\
\textbf{Our Method} & FT-FL & 65.39 & \textbf{64.15} & \textbf{63.42} & 65.06 & 63.98 & \textbf{64.40±0.72} \\
\bottomrule
\end{tabular}
}
\label{tab:auc_results}
\end{table}

We evaluate model performance using official metrics from the competition: Area Under the ROC Curve (AUC). Table~\ref{tab:auc_results} reports the results across five folds under different training strategies, including training from scratch (TFS), fine-tuning the rest of the network with the encoder frozen (FT-FL), and fine-tuning all parameters (FT-ALL).

Although our method does not achieve the highest AUC in every individual fold, it consistently ranks among the top in all folds and outperforms all baseline methods in terms of mean performance. Specifically, our method achieves the highest mean AUC (64.40) with the lowest standard deviations, demonstrating superior stability and generalization across folds.

Comparing the baselines, CNN-based models such as EfficientNet-B3-3D and ResNet50-3D trained from scratch show moderate performance but relatively high variance. Fine-tuning pretrained models (ResNet50-3D, SegResNet, and Swin-ViT3D) slightly improves individual results but does not consistently outperform training from scratch.

In contrast, methods integrating SAM-Brain3D as a feature extractor achieve notable improvements over traditional CNN and transformer-based models, including directly using a vanilla graph neural network as an adapter (SAM-Brain3D + GraphNet), replacing hypergraphs convolution networks with Graph Convolution Networks or Dynamic Hypergraph Neural Networks (denoted as SAM-Brain3D + HyDA-GCN and SAM-Brain3D + HyDA-DHGNN, respectively). This highlights the significance of leveraging high-quality, domain-specific pretrained feature extractors. Particularly, our method, based on SAM-Brain3D and HyDA, demonstrates the best overall generalization, achieving improvements in AUC while maintaining lower standard deviations.

These observations validate that SAM-Brain3D and HyDA significantly enhance feature representation quality, leading to more discriminative and robust embeddings for MGMT promoter status classification.

\section{Conclusion}
In this paper, we introduce a brain foundation model called SAM-Brain3D to segment diverse brain targets from 3D medical images, including up to 35 brain structures. We further propose a novel Hypergraph Dynamic Adapter (HyDA) to adapt SAM-Brain3D to downstream brain diagnosis tasks, e.g., Alzheimer's progression and MGMT classification. HyDA is designed to extract multi-modal information with hypergraphs and fuse multi-scale features with dynamic convolution, of which the kernels are dynamically generated and conditioned on each subject, facilitating subject-adaptive diagnosis. With SAM-Brain3D and HyDA, our method can achieve state-of-the-art performance on various segmentation and classification tasks for brain disease analysis. 

However, our method may have the following limitations. Firstly, we rely on complete modalities for training and inference on downstream tasks, but acquiring all the modalities (e.g., MRI, PET, and non-imaging data) can be expensive. As such, many subjects may have incomplete modalities and these subjects cannot be used for training or inference, limiting the flexibility of HyDA. Secondly, our method cannot effectively handle the class imbalance issue. This leads to our method excelling in some metrics like accuracy and specificity while failing in the others (e.g., AUC or sensitivity) as shown in our experimental results. Thus, how to effectively utilize incomplete modalities and better tackle the class imbalance issue can be interesting for future work.

\section*{Acknowledgment}
This project was supported with funding from the Cambridge Centre for Data-Driven Discovery and Accelerate Programme for Scientific Discovery, made possible by a donation from Schmidt Sciences. Zhongying Deng acknowledges the support from Wellcome Trust 221633/Z/20/Z. Zoe Kourtzi acknowledges support from the Biotechnology and Biological Sciences Research Council H012508 and BB/P021255/1, Alan Turing Institute TU/B/000095, Wellcome Trust 205067/Z/16/Z, 221633/Z/20/Z, Royal Society INF/R2/202107. Carola-Bibiane  Schönlieb acknowledges support from the Philip Leverhulme Prize, the Royal Society Wolfson Fellowship, the EPSRC advanced career fellowship EP/V029428/1, EPSRC grants EP/S026045/1 and EP/T003553/1, EP/N014588/1, EP/T017961/1, the Wellcome Innovator Awards 215733/Z/19/Z and 221633/Z/20/Z, the European Union Horizon 2020 research and innovation programme under the Marie Skodowska-Curie grant agreement No. 777826 NoMADS, the Cantab Capital Institute for the Mathematics of Information and the Alan Turing Institute. Angelica I. Aviles-Rivero gratefully acknowledges the support from Yau Mathematical Sciences Center, Tsinghua University. Chaoyu Liu acknowledges the support from the Maths4DL program under grant EP/V026259/1.











\bibliographystyle{elsarticle-num}
\bibliography{egbib}

\end{document}